# Machine Learning Performance Analysis to Predict Stroke Based on Imbalanced Medical Dataset

Yuru Jing*[a]
[a] University College London, Gower Street, London, UK, WC1E 6BT
* Author's e-mail: ucaby57@ucl.ac.uk

**ABSTRACT**

Cerebral stroke, the second most substantial cause of death universally, has been a primary public health concern over the last few years. With the help of machine learning techniques, early detection of various stroke alerts is accessible, which can efficiently prevent or diminish the stroke. Medical datasets, however, are frequently unbalanced in their class label, with a tendency to poorly predict minority classes. In this paper, the potential risk factors for stroke are investigated. Moreover, four distinctive approaches are applied to improve the classification of the minority class in the imbalanced stroke dataset, which are the ensemble weight voting classifier, the Synthetic Minority Over-sampling Technique (SMOTE), Principal Component Analysis with K-Means Clustering (PCA-Kmeans), Focal Loss with the Deep Neural Network (DNN) and compare their performance. Through the analysis results, SMOTE and PCA-Kmeans with DNN-Focal Loss work best for the limited size of a large severe imbalanced dataset (e.g., Stroke dataset), which is 2-4 times outperform Kaggle's work.

## 1. INTRODUCTION

According to the Global Burden of Disease (GBD) 2013 study, stroke is the world's second leading cause of mortality, accounting for approximately 11.8% which emerges when various fields in the brain are disrupted or diminished because of blood flow mechanism [1]. Therefore, it is essential to investigate the factors that may cause the stroke to reduce stroke morbidity and predict the stroke based on those attributes to intervention and treatment earlier to lower the mortality. A healthy lifestyle, which incorporates quitting smoking and drinking, controlling body mass index (BMI), average glucose level, and staying heart and kidney health, can help prevent stroke [1]. Furthermore, Machine learning is very vital and efficient in the decision-making processes of the prediction system, which has been successfully applied in both stroke prediction [1-2] and imbalanced medical datasets [3].

From 2007 to 2019, there were roughly 18 studies associated with stroke diagnosis in the subject of stroke prediction using machine learning in the ScienceDirect database [4]. Most researchers relied on more expensive CT/MRI data to identify the damaged area of the brain rather than using the low-cost physiological data [4]. Additionally, most earlier stroke predictions pertained to the complete and class balanced datasets. Support Vector Machine (SVM) is one of the most common traditional machine learning methods that frequently surpass the rest of other approaches [5-6], but recent research by Ge et al. implies that Deep Learning models are more feasible to attain the higher accuracy than classic machine learning techniques [7]. Nevertheless, few medical datasets are comprehensive and balanced; in fact, a large imbalanced or incomplete dataset ratio might lead to pointless or unreliable prediction [3]. As a result, prior works have used a variety of imputation approaches to cope with missing values, such as regression imputation and simple imputation [8], and some sophisticated work, such as Liu et al., will use Random Forest to pick features and impute simultaneously [3]. An imbalance in class distribution occurs when the minority classes are rare and the majority classes are prevalent, which can be alleviated from two dimensions, data or algorithm. At the data level, a major approach to rebalance data distribution is through resampling, in which undersampling the majority class or oversampling the minority class is used [3]. SMOTE is a well-known oversampling method that is generally effective and precise, albeit it runs the danger of overfitting, especially with large datasets [9]. Besides, clustering-based sampling is another powerful approach, which can be utilized for both undersampling and oversampling data [10-11]. At the algorithm level, modifying the algorithms can also enhance the performance of imbalanced datasets. Emon et al. implemented the ensemble weighted voting for ten baseline algorithms, resulting in a final prediction that beat the baseline algorithms alone [1]. In addition, in conventional prediction models, altering the loss function and the objective function is widespread [12-13]. Lin et al. revised the DNN-based loss function

(Focal Loss), focusing learning on difficult examples to fundamentally tackle the class imbalance issue straightforwardly and effectively [14]. A similar design was used in Liu et al.'s hybrid approach [3], which combines undersampling with a clustering method, as well as reweighting the loss function analogous to Focal Loss. Since it is usual for datasets in the medical field is imbalanced, we are supposed to pay more attention to it, especially for stroke prediction.

In this paper, I employed the low-cost physiological data, which has been overlooked in most previous studies but is also valuable and easy to obtain for stroke prediction. Missing value imputation and other methods such as SMOTE, Ensemble weighted voting, Focal Loss, and hybrid approach with Focal Loss (PCA-Kmeans with DNN Focal Loss) are supposed to be used before or combined with the machine learning algorithms to predict this incomplete and imbalanced stroke dataset successfully. The primary contribution of this work is as follows:

(1) Explore and compare influences of the different preprocessing techniques for stroke prediction according to machine learning. A regression imputation and a simple imputation are applied for the missing values in the stroke dataset, respectively. A label encoding, a one-hot encoding, and an ordinal encoding are used separately for the distinctive categorical features in the stroke dataset.

(2) For imbalanced classes, the performance of the Deep Neural Network (DNN) with Focal loss, the PCA-Kmeans with DNN-Focal Loss, the ensemble weight voting classifier, and the Synthetic Minority Oversampling Technique (SMOTE) application is compared to state-of-the-art classifiers such as DNN, Logistic Regression (LR), Stochastic Gradient Descent (SGD), Decision Tree Classifier (DTC), AdaBoost, Gaussian, Quadratic Discriminant Analysis (QDC), MultiLayer Perceptron (MLP), KNeighbors, Gradient Boosting Classifier (GBC), XGBoost (XGB).

## 2. METHODS

### 2.1 Feature Selection and Data Preprocessing

To obtain reliable prediction and advance computation speed from machine learning models, selecting proper features and preprocessing data are necessary procedures.

**Embedded feature selection - Random Forest:** As for the feature selection, there are primary three strategies to remove irrelevant and redundant features, which are filter, wrapper, and embedded strategy [15-16]. Liu et al. [16] proposed a Decision Tree (DTC) with a weighted Gini index embedded feature selection method that was excellent for imbalanced data classification, and its utility was confirmed by subsequent research experiments [17]. Besides, some statistical methods, such as correlation and variance inflation factors (VIF) among distinctive features, can also help feature selection.

**Data preprocessing:** As for processing data, outliers, duplicates, and missing values among the dataset can affect the performance of models [18]. Pursuant to standards from stroke attributes [19], outliers can be removed from the data. In this work, two imputation approaches are considered appropriate: simple imputation and regression imputation. A simple imputation replaces a missing value with a quantitative or qualitative attribute derived from other non-missing values, such as the median, mean, and mode. It is commonly used in most studies due to its simplicity, but it sometimes produces biased or unrealistic results. A regression imputation is also a preferred statistical technique for handling missing values. It firstly utilizes the complete features to construct the regression model and then exploits this regression model to fill the missing data. This method usually needs large samples of the complete features to generate consistent results, and it mainly works well for a single incomplete feature [20].

In the end, an embedded feature selection DTC, VIF, and correlation are used for feature selection, a simple imputation and a regression imputation are applied for dealing with missing values, standardization is employed for data scaling, and a label encoding, an ordinal encoding, and a one-hot encoding are utilized for categorical attributes transformation [21]. The results can be found in section 3.

### 2.2 Practical Approaches and Evaluation Metrics for Class Imbalance

**Synthetic Minority Over-sampling Technique (SMOTE):** SMOTE oversamples the minority class by taking each minority class sample and injecting synthetic examples along the line segments joining any/all the k minority class nearest neighbours, resulting in the minority class being more generic. Synthetic samples are made by multiplying the difference between the considered samples and their nearest neighbours by a random value between 0 and 1, then adding it to the considered samples [9].

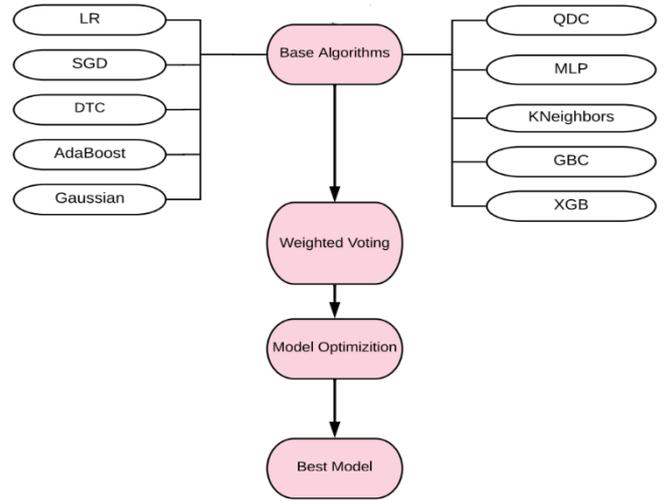

Figure 1. The procedure of Weighted Voting Classifier [1]

**Weighted Voting Classifier:** The weighted Voting Classifier is the approach to aggregating 10 base classifiers (LR, SGD, DTC, AdaBoost, Gaussian, QDA, MLP, KNeighbors, GBC, XGB) the above with the help of the weighted voting technique to approach the higher accuracy for prediction [1]. The procedure of this approach is shown in Figure 1.

**Deep Neural Network (DNN) with Focal Loss (FL):** Lin et al. [14] employed the focal loss to redefine the DNN-based loss function, which is intended to focus learning on hard examples to handle the class imbalance issue simply and effectively [3]. The focal loss reshapes the binary cross-entropy (CE) loss, down-weight majority examples, and concentrates on minority examples as follows:

$$CE(p, y) = CE(p_t) = -log(p_t) \quad (1)$$

$$p_t = \begin{cases} p, & y = 1 \\ 1 - p, & otherwise \end{cases} \quad (2)$$

$$FL(p_t) = -(1 - p_t)^\gamma \log(p_t), \quad \gamma \geq 0 \quad (3)$$

Where $p$ is the probability for the class label $y$, $(1 - p_t)^\gamma$ is a modulating factor of the binary CE loss with tunable focusing parameter.

**Principal Component Analysis with K-Means Clustering (PCA-Kmeans):** Principal Component Analysis (PCA) is a commonly used statistical technique for diminishing the dimension via transformation. I implement the PCA approach first to condense the original data set to a low-dimensional subspace, then clustered by one of the most efficient clustering algorithms, K-means [22]. Under this PCA-Kmeans process, undersampling is applied for the majority class of each cluster. To avoid the information loss and overlap of the majority class samples, instance selection will also be utilized in PCA-Kmeans, and the nearest $k_s$ samples are concerned as the majority class instances instead of the furthest samples [3].

**Evaluation Metrics:** Accuracy is widely used as one of the metrics in machine learning, but its less distinctiveness, less discriminability, less informativeness, and bias to majority class data [23]. Therefore, it will also be considered in this work but will not be the most crucial metric regardless of binary classification or imbalanced classification. Since F1-score and Area under the ROC Curve (AUC) are also found to be good discriminators, outperforming accuracy in classifier optimization for binary classification issues, I will be picked AUC and F1-score as the most vital measure of classification models in this work [23].

## 3. EXPERIMENTS AND RESULTS

### 3.1 Data Description

In this study, the dataset of the stroke is derived from the Kaggle competition with details listed as Table 1. The dataset is typically an imbalanced class set containing 11 input features and 1 target, stroke. The redundant feature (patient id) and outliers in BMI (outlier: BMI > 60%), Age (outlier: Age < 25), and" Other" gender (with only 1 sample) based on stroke standards need to remove from the original data set [17]. What is more, it is found that the missing values only occur in BMI, so a simple imputation and a regression imputation will also be used here. Meanwhile, ordinal encoding is applied to smoking status because of the ordinal characteristics of the smoking degree. The rest of the categorical attributes will be transformed to numerical by either label encoding or one-hot encoding.

Table 1. Dataset Description.

| Features | Values | Features | Values |
| --- | --- | --- | --- |
| Patient ID | 1-5110 | Hypertension | Yes/No |
| Gender | Male/Female/Other | Ever married | Yes/No |
| Residence type | Rural/Urban | Age | 0.08-82 |
| Avg-glucose | 55-272 | Heart disease | Yes/No |
| Work type | Self-employed/Private/Government | BMI | 10-98 |
| Smoking Status | Unknown/Never/Formerly/Smoked | | |

### 3.2 Experiment Setup

After cleaning and processing the data set, there are only 3806 samples in total, and less than 5% samples are the stroke. The combination of two imputations (simple or regression) and two encodings (label or one-hot) will produce four different transformed data sets. 10 state-of-art machine learning algorithms the above and simple DNN with three hidden layers and ReLu functions will be implemented as the baseline, and it is compared to four types of approaches for imbalanced classification the above (SMOTE, Weighted Voting Classifier, DNN with Focal Loss, undersampling majority class by PCA-Kmeans with DNN-Focal Loss) with 4 distinctive transformed data (simple/regression imputation, label/0ne-hot encoding).

### 3.3 Performance Analysis and Results

#### 3.3.1 Feature Selection

Based on the feature importance from DTC in Figure 2, the least vital features are heart disease and ever married. However, the previous medical analysis mentioned there appeared to suggest a relationship between stroke and heart disease [3] and Figure 3 below also shows that heart disease relatively strongly correlates with gender, age, average glucose level, and stroke when compared to the overall correlation analysis. VIF analysis in table 2 also indicates that no features are excessively collinear. As a result, all features are chosen based on the feature selection.

Table 2. The VIF of mean/regression imputation with label/one-hot encoding in work type are nearly the same.

| Features | Mean/Regression Label | Mean/Regression One-Hot |
| --- | --- | --- |
| Gender | 1.3 | 2.5 |
| Age | 1.1 | 7.0 |
| Hypertension | 1.0 | 3.0 |
| Heart Disease | 1.7 | 1.4 |
| Ever Married | 1.2 | 1.1 |
| Work Type (all) | 1.2 | (Self) 1.0, (Private) 1.0, (Govern) 1.1 |
| Residence Type | 3.8 | 1.1 |
| Avg-glucose | 2.9 | 1.1 |
| BMI | 1.9 | 1.0 |
| Smoking Status | 1.9 | 1.0 |
| Stroke | 1.1 | 1.1 |

#### 3.3.2 Stroke Factors Analysis

To figure out if some potential factors affect the stroke, I visualize the cleaned data set by categories, and the results are shown below. The overall distributions and regression relations for all numerical features, age, average level glucose, and BMI are as shown in Figure 4. Moreover, the further stratified comparison of distinctive features categorized by stroke. It has been discovered that although there are no significant differences between the shape of distributions from three numerical features, the overall average glucose value of stroke patients is greater than that of healthy individuals, and the age of stroke patients is centralized on older people. The potential categorical features that may lead to stroke are hypertension and heart disease.

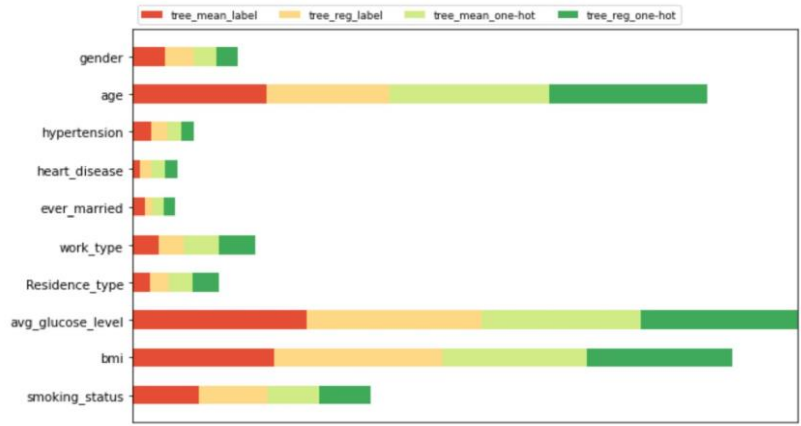

Figure 2. Feature importance based on an embedded method, decision tree. Tree mean label, tree reg label, tree mean one hot, tree reg one hot respectively indicates 4 different data set the above (mean imputation with label encoding, regression imputation with regression label, mean imputation with one hot encoding, regression imputation with one hot regression).

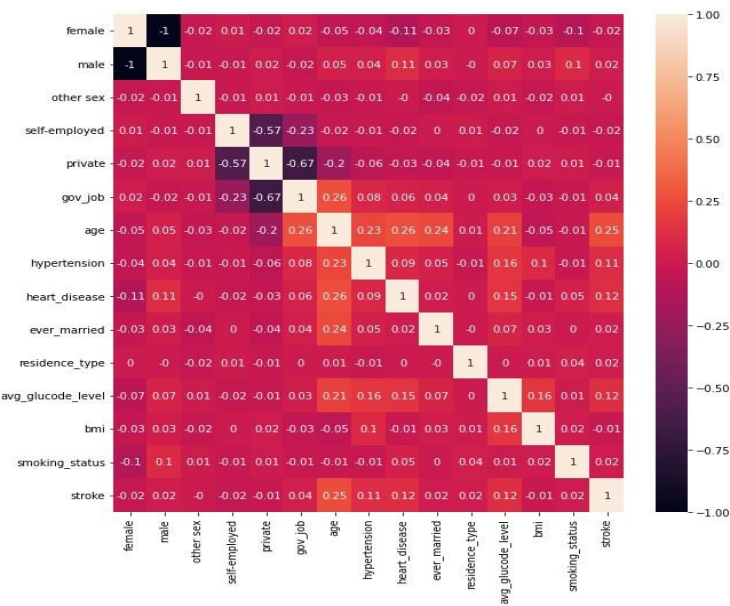

Figure 3. The correlation analysis for regression imputation one hot encoding data set.

### 3.3.3 Imbalanced Class (Stroke) Approaches Comparison

Repeating experiments of the above 4 datasets are found that a regression imputation is the same or slightly better than a mean imputation (0-2% accuracy or 0.00-0.03 outperform), and there is no significant difference between label encoding and one-hot encoding in this work. Hence, the regression imputation with one-hot encoding will be used and shown in the following experiments. Since model metrics for the test set are the most important ones that should be considered, all final metric values below are related to the test set. Table 3 shows the comparison results and parameters of ten baseline algorithms with a weighted voting classifier and SMOTE. The DNN only and DNN with Focal Loss and SMOTE DNN with Focal Loss and PCA-Kmeans with DNN-Focal Loss' comparison results and the details of simple DNN layers and training are indicated in table 4 below.

For stroke classification, Weighted voting performed slightly better than the majority of the other baseline algorithms but performed worse than QDA and DTC baseline in terms of AUC. SMOTE works very well for the imbalanced class, which beats all baseline models and weighted voting. Compared to DNN alone and the other 3 techniques, it is evident that focal loss can improve the classification of an imbalanced dataset but may not for a balanced dataset.

In the case of PCA-Kmeans with DNN-Focal, a unique design comparable to Liu et al.'s study, I first construct the new transformed dataset employing variance by explained and" Elbow method" [24], as illustrated in Figure 5. Then, I select the hyperparameter, $k_s$ with the restricted scope, 3,5,7,9,11, and can decrease the classification error rate and imbalance ratio [3]. Since there are 5 clusters is generated from PCA-KMeans, I preserve all minority class (stroke) samples and combine them with the extreme cases, the nearest neighbours majority class(non-stroke) samples with the sizes 3, 5, 7, 9, 11, respectively in each cluster, and this 5 clusters will become 5 batches for the later DNN-Focal Loss training. Through this undersampling instance selection, the new dataset size now is approximately 2 times greater than the original data size, and the imbalance ratio is reduced by around 2 times, which has a similar effect as data augmentation, with the result being far superior to the rest methods. Meanwhile, as the increasing of epochs, the metrics F1 AUC) of classification prediction for imbalanced data without Focal Loss will decrease, while the imbalanced data with Focal Loss will increase.

In terms of current work, the best model without the SMOTE technique is QDA, which is comparable to most SMOTE-enabled algorithms. Regarding the application of the SMOTE technique, the best models are SGD, DTC, and LR, with the classification metric, F1 score, and AUC approximately 0.30 and 0.71, respectively, and the results are stable even after tuning the parameters and validated automatically with the technique of Grid Search Cross-Validation (CV) [25]. In my novel design, similar to Liu et al. 's work [3] for simple DNN, applying PCA-Kmeans for undersampling majority class through instances selection to the simple DNN with Focal Loss works best, outperforming most previous studies by 2-4 times as epochs rise.

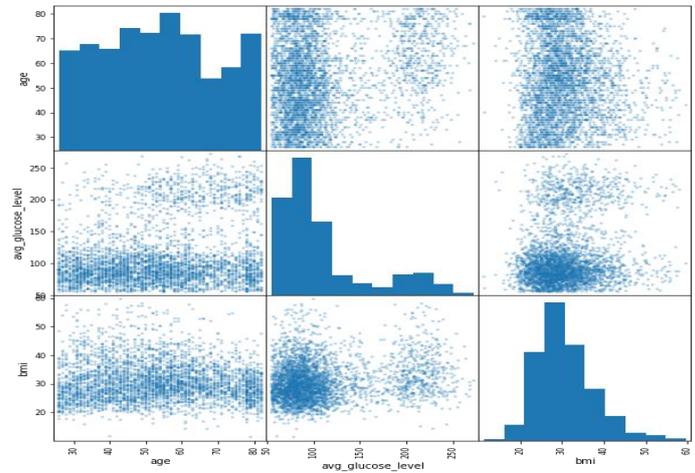

Figure 4. The histogram and scatter plotting matrix for all numerical features, age, average glucose level, and bmi.

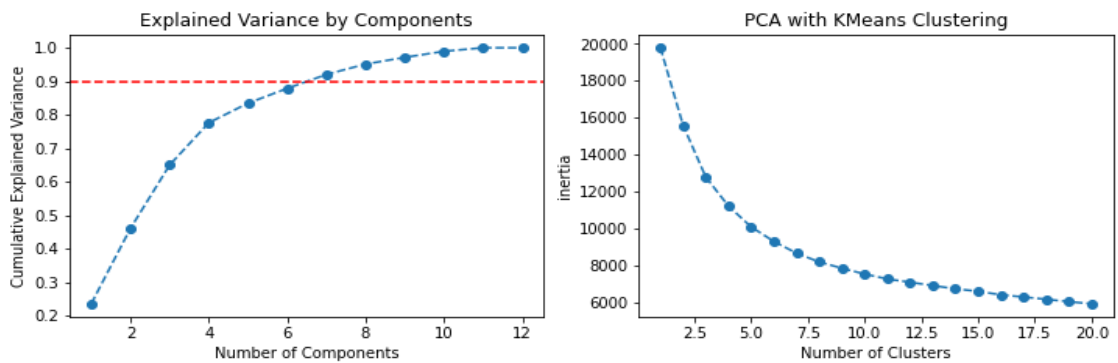

Figure 5. The left graph is the explained variance by components, more than 90% data is illustrated when 7 components are selected. The right graph is PCA-Kmeans Clustering, 5 clusters are chosen based on the slope.

Table 3. Comparison of the baseline and SMOTE with baseline Metrics (F1 and AUC) in test dataset. (Split 20% of the original dataset as the test set.) Below are the best parameters after tuning and default is the default setting in the sklearn.: LR('l1', C=0.2336, solver='liblinear'), SGD(loss='hinge', 'l2'), DTC('entropy', max_depth=3), AdaBoost(learning_rate=0.5, n_estimators=200), Gaussian(default), QDA(reg_param=0.1, store_covariance=True, tol=0.0001), MLP(default), KNeighbors(leaf_size=1, n_neighbors=11, p=1), GBC(n_estimators=80), XGB(default).

| Classifier | F1 | AUC | Accuracy | SMOTE+Classifier | F1 | AUC | Accuracy |
|---|---|---|---|---|---|---|---|
| LR | 0.00 | 0.50 | 92% | **SMOTE + LR** | **0.28** | **0.71** | 70% |
| SGD | 0.00 | 0.50 | 92% | **SMOTE + SGD** | **0.30** | **0.73** | 67% |
| DTC | 0.19 | 0.57 | 87% | **SMOTE + DTC** | **0.28** | **0.71** | 69% |
| AdaBoost | 0.02 | 0.51 | 92% | SMOTE + AdaBoost | 0.24 | 0.61 | 81% |
| Gaussian | 0.00 | 0.50 | 92% | SMOTE + Gaussian | 0.23 | 0.60 | 78% |
| **QDA** | **0.25** | **0.61** | 85% | SMOTE + QDA | 0.27 | 0.71 | 67% |
| MLP | 0.08 | 0.52 | 92% | SMOTE + MLP | 0.24 | 0.60 | 85% |
| KNeighbors | 0.06 | 0.51 | 92% | SMOTE + KNeighbors | 0.22 | 0.60 | 76% |
| GBC | 0.00 | 0.50 | 91% | SMOTE + GBC | 0.18 | 0.55 | 86% |
| XGB | 0.10 | 0.52 | 91% | SMOTE + XGB | 0.11 | 0.53 | 89% |
| Weighted Voting | 0.09 | 0.54 | 91% | SMOTE + Weighted Voting | 0.27 | 0.64 | 80% |

Table 4. Comparison of DNN only and DNN with Focal Loss, SMOTE DNN with Focal Loss, and PCA-Kmeans DNN-Focal Loss with Metrics (F1 and AUC) in test dataset. (Split 20% of the original dataset and new dataset after PCA-Kmeans as the test set. 50, 100, and 200 epochs are also considered to make the comparison. DNN is fully connected (out channel is 64, 128, 256, with ReLU activation after each layer, the last layer is 1 out channel linear layer with Sigmoid. Threshold is 0.25, when final output > 0.25, the test predict label is 1 (stroke), otherwise is 0 (non-stroke)).

| Model | Epoch | F1 | AUC | Accuracy |
|---|---|---|---|---|
| DNN | 50 | 0.22 | 0.57 | 89% |
|  | 100 | 0.17 | 0.55 | 88% |
|  | 200 | 0.15 | 0.53 | 86% |
| **DNN + Focal Loss** | 50 | 0.25 | 0.62 | 78% |
|  | 100 | 0.29 | 0.68 | 76% |
|  | **200** | **0.31** | **0.72** | 75% |
| DNN + SMOTE + Focal Loss | 50 | 0.24 | 0.58 | 80% |
|  | 100 | 0.26 | 0.59 | 78% |
|  | 200 | 0.28 | 0.64 | 75% |
| **PCA-Kmeans + DNN + Focal Loss** | 50 | 0.33 | 0.61 | 86% |
|  | 100 | 0.47 | 0.76 | 71% |
|  | **200** | **0.77** | **0.90** | **92%** |

## 4. DISCUSSION

SMOTE is an excellent technique for the imbalanced class dataset, and it can be utilized with other machine learning algorithms and advance them a lot. Without it, only QDA has some resistance to the risk of minority class misprediction in an imbalanced dataset. Next, Weighted Voting works not so well as the Emon [1] mentioned, which may be because of the implementation of the larger and balanced dataset and less focus on outliers of stroke standards in Emon's research. Thirdly, the novel design, PCA-Kmeans with DNN Focal Loss, has a significant impact on data augmentation, as it not only reduces the imbalance ratio of the original dataset but also enhances its size. Fourthly, the regression imputation has a slightly better effect on the classification than a simple imputation, which may provide the probabilities that selecting the proper imputation of the missing values can also help the development of the models. Last but not least, while SMOTE is a great technique for imbalanced classes, it also has the risk of increasing overlapping and extra noises, especially for high-dimensional data [9].

## 5. CONCLUSION

The paper investigates and compares four methods for resolving misclassification in the minority class in an extremely unbalanced stroke dataset. In terms of baseline algorithms, only QDA can withstand the imbalanced impact to some extent, as it has nearly the same F1 score as the rest of the algorithms using the SMOTE technique. All machine learning algorithms benefit greatly from SMOTE, and LR, SGD, and DTC work best with SMOTE. Compared to the other two algorithms mentioned above, LR is preferable due to its simplicity and efficiency, especially when working in a clinical setting. Although DNN with Focal Loss works nearly as well as LR with SMOTE for imbalanced classes, the novel design, PCA-KMeans with DNN Focal Loss, performs 2-4 times better than all other models. This design can be used in conjunction with Liu et al.'s work [3], because their designs are comparable, and the modified loss function in Liu's paper is similar to Focal Loss as well, and it still can strengthen classification for imbalanced datasets when the number of epochs or the dataset size increased.

## ACKNOWLEDGMENTS

Thank Dr.Petru Manescu in UCL so much for providing primary paper resources on the approaches for imbalanced dataset classification and the UCL CS department faculty for comments that improved this paper.

## REFERENCES

[1] Emon, M. U., Keya, M. S., Meghla, T. I., Rahman, M. M., Mamun, M. S. A., and Kaiser, M. S., "Performance Analysis of Machine Learning Approaches in Stroke Prediction," *Proceedings of the 4th International Conference on Electronics, Communication and Aerospace Technology, ICECA 2020* , 6 (2020).

[2] Amini, L., Azarpazhouh, R., Farzadfar, M. T., Mousavi, S. A., Jazaieri, F., Khorvash, F., Norouzi, R., and Toghianifar, N., "Prediction and control of stroke by data mining," *International Journal of Preventive Medicine* **4**, S245–S249 (2013).

[3] Liu, T., Fan, W., and Wu, C., "A hybrid machine learning approach to cerebral stroke prediction based on imbalanced medical dataset," *Artificial Intelligence in Medicine* **101**(August), 9 (2019).

[4] Sirsat, M. S., Ferm´e, E., and Cˆamara, J., "Machine learning for brain stroke: a review," *Journal of Stroke and Cerebrovascular Diseases* **29**(10), 105162 (2020).

[5] Bento, M., Souza, R., Salluzzi, M., Rittner, L., Zhang, Y., and Frayne, R., "Automatic identification of atherosclerosis subjects in a heterogeneous mr brain imaging data set," *Magnetic resonance imaging* **62**, 18–27 (2019).

[6] Rebouc¸as Filho, P. P., Sarmento, R. M., Holanda, G. B., and de Alencar Lima, D., "New approach to detect and classify stroke in skull ct images via analysis of brain tissue densities," *Computer methods and programs in biomedicine* **148**, 27–43 (2017).

[7] Ge, Y., Wang, Q., Wang, L., Wu, H., Peng, C., Wang, J., Xu, Y., Xiong, G., Zhang, Y., and Yi, Y., "Predicting post-stroke pneumonia using deep neural network approaches," *International journal of medical informatics* **132**, 103986 (2019).

[8] Thabet, A., "Enhanced Robust Association Rules (ERAR)Method for Missing Values Imputation," *International Journal of Advanced Trends in Computer Science and Engineering* **9**(4), 6036–6042 (2020).

[9] Chawla, N. V., Bowyer, K. W., Hall, L. O., and Kegelmeyer, W. P., "SMOTE: Synthetic Minority Oversampling Technique," *Journal of Artificial Intelligence Research* **16**(Sept. 28), 321–357 (2002).

[10] Zhang, J., Wang, T., Ng, W. W., Zhang, S., and Nugent, C. D., "Undersampling near decision boundary for imbalance problems," in [*2019 International conference on machine learning and cybernetics (ICMLC)*], 1–8, IEEE (2019).

[11] Liang, X., Jiang, A., Li, T., Xue, Y., and Wang, G., "Lr-smote—an improved unbalanced data set oversampling based on k-means and svm," *Knowledge-Based Systems* **196**, 105845 (2020).
[12] Wang, Y., Yang, L., and Ren, Q., "A robust classification framework with mixture correntropy," *Information Sciences* **491**, 306–318 (2019).
[13] Ghazikhani, A., Monsefi, R., and Sadoghi Yazdi, H., "Online cost-sensitive neural network classifiers for non-stationary and imbalanced data streams," *Neural computing and applications* **23**(5), 1283–1295 (2013).
[14] Tsung-Yi Lin, "Focal Loss for Dense Object Detection (RetinaNet)," *13C-NMR of Natural Products* , 30–33 (2017).
[15] Khalid, S., Khalil, T., and Nasreen, S., "A survey of feature selection and feature extraction techniques in machine learning," *Proceedings of 2014 Science and Information Conference, SAI 2014* , 372–378 (2014).
[16] Liu, H., Zhou, M., and Liu, Q., "An embedded feature selection method for imbalanced data classification," *IEEE/CAA Journal of Automatica Sinica* **6**(3), 703–715 (2019).
[17] Chen, R.-C., Dewi, C., Huang, S.-W., and Caraka, R. E., "Selecting critical features for data classification based on machine learning methods,"
[18] Ben-Gal, I., "Outlier detection," in [*Data mining and knowledge discovery handbook*], 131–146, Springer (2005).
[19] Liu, M., Wu, B., Wang, W.-Z., Lee, L.-M., Zhang, S.-H., and Kong, L.-Z., "Stroke in china: epidemiology, prevention, and management strategies," *The Lancet Neurology* **6**(5), 456–464 (2007).
[20] Emmanuel, T., Maupong, T., Mpoeleng, D., Semong, T., Mphago, B., and Tabona, O., [*A survey on missing data in machine learning*], vol. 8, Springer International Publishing (2021).
[21] Hancock, J. T., Khoshgoftaar, T. M., and Hancock, K. J., "Survey on categorical data for neural networks," *Big Data* **7**, 28 (2020).
[22] Ding, C. and He, X., "K-means clustering via principal component analysis," in [*Proceedings of the twentyfirst international conference on Machine learning*], 29 (2004).
[23] M, H. and M.N, S., "A Review on Evaluation Metrics for Data Classification Evaluations," *International Journal of Data Mining Knowledge Management Process* **5**(2), 01–11 (2015).
[24] Marutho, D., Handaka, S. H., Wijaya, E., et al., "The determination of cluster number at k-mean using elbow method and purity evaluation on headline news," in [*2018 international seminar on application for technology of information and communication*], 533–538, IEEE (2018).
[25] Soper, D. S., "Greed is good: Rapid hyperparameter optimization and model selection using greedy k-fold cross validation," *Electronics (Switzerland)* **10**(16) (2021).